\pdfoutput=1

\documentclass[11pt]{article}

\usepackage[final]{acl}
\usepackage{algpseudocode}
\usepackage{times}
\usepackage{colortbl}
\usepackage{xcolor}

\definecolor{mycustomcolor}{HTML}{daebfa}
\usepackage{latexsym}
\usepackage{amsmath}
\usepackage{amssymb} 
\usepackage{booktabs}
\usepackage{multirow}
\usepackage{graphicx}
\usepackage{xspace}
\usepackage{tcolorbox}
\usepackage[T1]{fontenc}
\usepackage{amstext}
\usepackage{varwidth}
\usepackage{tcolorbox}
\usepackage{soul}
\newcommand{\datasetFont}{\textbf}
\newcommand{\ours}{\datasetFont{D-{UE}}\xspace}

\usepackage[utf8]{inputenc}

\usepackage{microtype}

\usepackage{inconsolata}

\usepackage{graphicx}

\title{LLM Uncertainty Quantification through Directional Entailment Graph and Claim Level Response Augmentation}

\author{$\text{Longchao Da}^{1}$, $\text{Tiejin Chen}^{1}$, $\text{Lu Cheng}^{2}$, $\text{Hua Wei}^{1}\thanks{\ \ Corresponding author.}$ \\
  $\text{}^{1}$Arizona State University, $\text{}^{2}$University of Illinois Chicago \\
  \texttt{\{longchao, tchen169, hua.wei\}@asu.edu, lucheng@uic.edu} \\
  }

\begin{document}
\maketitle
\begin{abstract}
The Large language models (LLMs) have showcased superior capabilities in sophisticated tasks across various domains, stemming from basic question-answer (QA), they are nowadays used as decision assistants or explainers for unfamiliar content. However, they are not always correct due to the data sparsity in specific domain corpus, or the model's hallucination problems. Given this, how much should we trust the responses from LLMs? This paper presents a novel way to evaluate the uncertainty that captures the directional instability, by constructing a directional graph from entailment probabilities, and we innovatively conduct Random Walk Laplacian given the asymmetric property of a constructed directed graph, then the uncertainty is aggregated by the derived eigenvalues from the Laplacian process. We also provide a way to incorporate the existing work's semantics uncertainty with our proposed layer. Besides, this paper identifies the vagueness issues in the raw response set and proposes an augmentation approach to mitigate such a problem, we conducted extensive empirical experiments and demonstrated the superiority of our proposed solutions. 
\end{abstract}

\section{Introduction}
The Large Language Models (LLMs)~\cite{chang2024survey} have become a hot spot for almost everyone, they demonstrate superior performance on various tasks and even proved to be able to conduct human-like conversations by breaking the Turing Test~\cite{biever2023chatgpt}. There are different voices on the emerging LLM techniques, and there are also different attitudes towards it~\cite{kambhampati2024llms, valmeekam2022large}, either skeptical or accepting, one major concern is commonly acknowledged that the trustworthiness of LLMs responses is not guaranteed~\cite{sun2024trustllm, liu2023trustworthy, huang2024survey}. The trustworthiness has become a key obstacle for LLMs to deploy in crucial scenarios, such as healthcare~\cite{yang2023large, wang2024enhancing}, autonomous control~\cite{wang2024survey}, and intelligent planning~\cite{kambhampati2024llms, da2024open}. 

This has brought many researchers to investigate the uncertainty quantification (UQ) approaches to better understand the LLM's inferences, e.g., how well they estimate the system dynamics given some context information~\cite{da2024prompt}. However, the UQ in Natural Language Generation (NLG), brings distinct challenges by their intrinsic semantics features, linguistic ambiguity, and complex output structures~\cite{lin2023generating}. 
Another challenge in UQ for LLMs lies in the limited access to commercial large models, the unavailable model parameters or true prediction probabilities greatly hinder the intrinsic profiling of the model's behavior, and leading to unachievable white-box uncertainty evaluation~\cite{balloccu2024leak}. 
\begin{figure}[t!]
    \centering
    \includegraphics[width=0.5\textwidth]{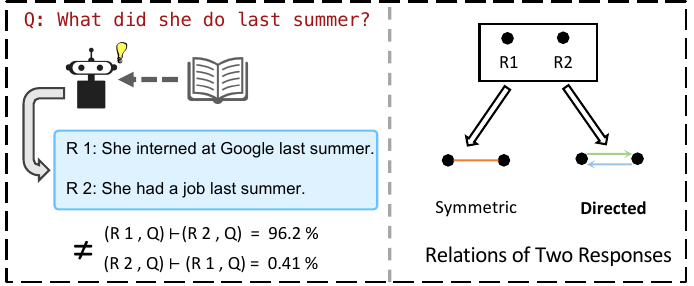}
    \caption{The left part is an example of directional entailment logic, (R1, Q) $\vdash$ (R2, Q) means the probability of R1 entails R2 given the context of question Q, and the right part shows the difference between existing symmetric similarity and our proposed directed relations.}
    \label{fig:demo1}

\end{figure}

\begin{figure*}[h!]
    \centering
    \includegraphics[width=0.99\textwidth]{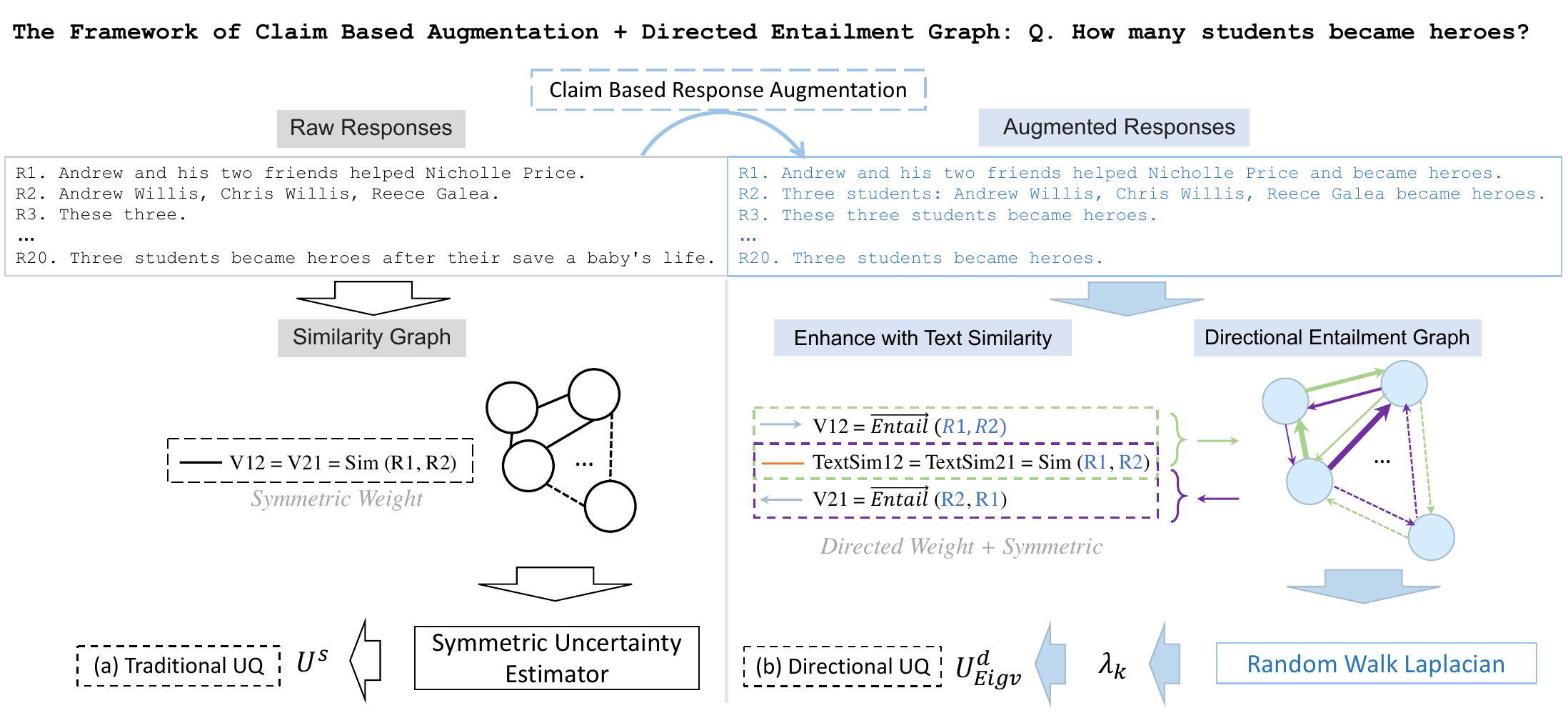}
    \caption{The overall directional uncertainty quantification (UQ) framework of \ours (right) compared to the traditional symmetric similarity-based uncertainty evaluation (left). As shown in the figure, the traditional method uses symmetric-based similarity and feeds into an estimator (e.g., Numset, Symmetric Laplacian, etc.) that only perceives monotonous semantics uncertainty $U^{s}$, while \ours perceives both directions of entailment between response pairs and enhanced by text similarity, the Random Walk Laplacian is specially applied for complex and asymmetric property. Specifically, after Random Walk Laplacian, we derive the eigenvalues $\lambda_k$ from Laplacian and aggregate them following Eq. ~\ref{eq:egen} as the final uncertainty measurement $U^{d}_{Eigv}$. We also provide a way to fairly consider both semantic uncertainty and directional uncertainty in Section~\ref{sec:agg}.}
    \label{fig:mainfig}
\end{figure*}
Alternatively, researchers resort to the black-box quantification~\cite{lin2023generating} within limited question and response sets. A common practice for black-box evaluation is to first build a similarity matrix from a set of responses, and detect the inconsistency of these responses by either conducting Graph Laplacian or analyzing the response set's entropy~\cite{kuhn2023semantic}. However, the current works of literature only consider `how similar' are the two responses when constructing the matrix, this assumes the similarity from response A to B is the same as B to A. But in fact, in linguistics studies, two sentences contain directional logic information. For example, as in Fig.~\ref{fig:demo1}, this pair of responses contains two dramatically different entailment probabilities (measured by the NLI model~\footnote{\url{https://huggingface.co/microsoft/deberta-v3-large}}). This implies potential direction information in the response set that existing work neglects by taking the mean or based on semantic measures (undirected). 

In this paper, we propose a novel way named \underline{\textbf{D}}irected \underline{\textbf{U}}ncertainty \underline{\textbf{E}}valuation \ours to apply a directed graph enforced by entailment probability to construct a more nuanced relationship that can capture the directions of responses and carry the semantic similarities at the same time. Besides, we also discover that the generated responses themselves may have vagueness issues that, bring more challenges to the UQ process, in this paper, we also propose a claim-based augmentation method that helps reduce the vagueness issue and mine the model's real awareness of a question, further enhanced the UQ for LLMs. 

\section{Related Work}

The first branch of research is solving the UQ in LLMs by inducing the LLMs to output their uncertainty along with the response~\cite{kadavath2022language, lin2022teaching, mielke2020linguistic, tian2023just}. Most of the literature above requires the token-level probabilities of LLMs to train (or fine-tune) and predict the uncertainty. This is a straightforward solution while having full access to model structure and weights, but it can be unwieldy as it is time-consuming and resource-tense. Another method~\cite{kuhn2023semantic} estimates LLMs uncertainty directly from response level semantic entropy,  yet still requires the token-related probability values as input, which is hard to access given black-box or commercial language models. 

In consideration of fast and lightweight evaluation, some researchers propose to solve the UQ by treating the LLMs as black-box and analyzing the consistency in the response semantics structure.~\cite{lin2023generating} first analyzes the UQ by text responses, treating the sum of eigenvalues from the graph Laplacian as the uncertainty indicator.~\cite{chen2023quantifying} identify unreliable or speculative answers by computing a confidence score for its generated outputs. However, they solely analyze the UQ from semantics, and ~\cite{lin2023generating} take the average of entailment probability from two directions to construct a similarity matrix, while in this paper, we find the claims together with the semantics information, better contribute to more comprehensive uncertainty quantification, and the directional logic, is not negligible in nuance analysis of response intrinsic structures.

\section{Preliminaries}
In this section, we will formalize the uncertainty evaluation in LLMs. 
Let $\mathcal{M}$ be a general LLM model, which is trained from a certain network structure and contains a parameter set $\theta$. In the prompt-based inference period, an input $x$ is provided to $\mathcal{M}$ and the model produces a sequence of tokens denoted as $\hat{y}$ from a probability distribution $ \mathbf{p} (\hat{y}|x, \theta)$. The probability distribution plays a key role in understanding the $\mathcal{M}$'s characteristics. 

For those who train the model from scratch, or use fully open-sourced models, the probability logits (or even the model parameters) are available for analysis and evaluation, and this branch of practice is seen as \textbf{White Box} evaluation, while on the other hand, due to the commercial, competitive, or other reasons, if there is no direct access to the probability logits, the uncertainty evaluation under this scenario is taken as \textbf{Black Box} evaluation.  

\subsection{White Box Evaluation}
Traditionally, researchers could conduct uncertainty evaluation by gradient norms either from the input aspect or parameter aspect: 
$
U_{\text{grad-input}}(x) = \left\| \nabla_x \mathcal{L}( \hat{y}, y ) \right\|_2
$ 
$
U_{\text{grad-param}}(x) = \left\| \nabla_\theta \mathcal{L}( \hat{y}, y ) \right\|_2
$, where $\mathcal{L(\cdot|\cdot)}$ is the loss function between predicted output $\hat{y}$ and groundtruth $y$. And the gradient is calculated from two different aspects to understand the model's sensitivity. This requires the true label of answer $y$, which is not suitable for open-ended questions or prompts. An alternative solution is to evaluate the inconsistency based on the list of responses $\textbf{Y} = \{y_1, y_2, y_3, ... y_n\}$, and their probabilities $\textbf{P}$. Such as the use of entropy for inconsistency implementation: 
    $U(x) = H(\textbf{Y}|x) = - \sum_{y} p(\textbf{y}|x) log(p(\textbf{y}|x))$
where the $x$ is the input and $\textbf{Y}$ is the sequence of generated tokens (as a response). ~\cite{kuhn2023semantic, sun2019feature, abdar2021review}. 

\subsection{Black Box Evaluation}
For black box evaluation, the evaluator only has the text-level responses~\cite{wang2023augmenting}, this typically requires a more nuanced and deeper understanding of the model's output stability and potential response structure from limited Q-A samples. Assume there are $\textit{\textbf{n}}$ response samples to the same question $\textit{\textbf{q}}$. The common practice is to construct a matrix $\mathcal{S}$ which encapsulates the similarity information among the responses: 
\begin{equation}
    \mathcal{S} = \begin{bmatrix}
    1 & s_{12} & s_{13} & \cdots & s_{1n} \\
    s_{21} & 1 & s_{23} & \cdots & s_{2n} \\
    s_{31} & s_{32} & 1 & \cdots & s_{3n} \\
    \vdots & \vdots & \vdots & \ddots & \vdots \\
    s_{n1} & s_{n2} & s_{n3} & \cdots & 1
\end{bmatrix}
\end{equation}

where each of the value at position $s_{ij}$, \{$i, j \in (1 \sim n)$\} is the calculated pariwise similarity score.

Given a list of responses \( R = \{r_1, r_2, \ldots, r_n\} \), the pairwise similarity score \( s_{ij} \) between responses \( r_i \) and \( r_j \) can be calculated using a general similarity function as:
$s_{ij} = \text{sim}(r_i, r_j)$
the current explorations on matrix $\mathcal{S}$ are mainly based on symmetric similarity property calculations such as Jaccard similarity or worldVector similarity, implying the $s_{ij} = s_{ji}$ in $\mathcal{S}$. Then this condition guarantees the use of Normalized Laplacian to understand the hidden structure in the responses space~\cite{lin2023generating}:
\begin{equation}
L := I - D^{-\frac{1}{2}} W D^{-\frac{1}{2}}
\end{equation}
where the weighted adjacency matrix $W$ is from the symmetric similarity matrix $\mathcal{S}$, and the degree matrix is: 
\begin{equation}
D_{r_i, r_j} = 
\begin{cases} 
\sum_{j' \in [n]} w_{i, j'} & (r_i = r_j) \\
0 & (r_i \neq r_j)
\end{cases}
\end{equation}
where the diagonal element \( D_{r_i,r_j} \) ($ r_i = r_j$) is the degree of the node \( r_i \), which is the sum of the weights (similarity $s_{i, j'}$) of all edges connected, $j'$ goes over the connected responses with size $n$.

Then one can leverage the constructed Symmetric Laplacian to find the eigenvalue to represent the connectivity of the graph, and use this as an indicator of uncertainty:
$U_{\text{EigV}} = \sum_{k=1}^{n} \max(0, 1 - \lambda_k)$
where the $\lambda_k$ is the $k_{\text{th}}$ eigenvalues of Laplacian $L$.

\subsection{Discussion}\label{sec:discuss}
Since the white box evaluation places a strict requirement on the original model, we analyze it from the perspective of the black box evaluation in this paper. 

\textbf{First}, the current black box evaluation makes assumption that $s_{ij} = s_{ji}$, however, in the actual knowledge representation logic, this neglected the directional information of two responses: 
If proposition \( A \) entails proposition \( B \) (denoted as \( A \vdash B \)), it means that if \( A \) is true, then \( B \) must also be true. This is a one-way relationship. Importantly, this relationship is not necessarily symmetric; that is, \( A \vdash B \) does not imply \( B \vdash A \). So the construction of a symmetric matrix broke this rule, which will lead to the loss of directional information from the original response set. In this paper, we propose to reconstruct the response relationship from a directional graph and provide a Random Walk Laplacian uncertainty evaluation method to better fit the asymmetric property of the constructed graph. 



\textbf{Second}, the response set with long answers, containing more than one identical claim is easy to be miscalculated on the similarity from either semantics or knowledge claim aspect.  E.g, to the question \textcolor{blue}{`How many students became heroes?'} the two answers from language model $\mathcal{M}$ as: \textcolor{orange}{A: `Andrew Willis, Chris Willis, Reece Galea'}  and \textcolor{orange}{B: `Three students became heroes'}. According to the context, the answer A is partially correct because it named the correct persons in the answer, however, the similarity between A and B is near 0 either calculated from entailment similarity or Jaccard, etc.  This raises our proposal to provide claim-based augmentation before the uncertainty evaluation to recover the correct response intentions.

\section{Uncertainty Evaluation within Directed Entailment Graph
: \ours}

In this section, we will discuss how to formally model the logical direction information~\cite{kripke1959distinguished, dagan2004probabilistic} in the responses with different entailment probabilities, and how the claims-based response augmentation helps with the potential semantic information mining. And then, we provide a way to integrate our method with plain semantic similarity matrix-derived uncertainty, which makes our method possible to layer on any of the existing methods that overlook the directional entailment information. The overall framework of \ours compared to the traditional UQ based on the symmetric measure is shown in Fig.~\ref{fig:mainfig}.

\subsection{Directional Entailment Graph}

In order to preserve the directional entailment information from a response set $R$, we adopt the NLI (Natural Language Inference) model to provide pair-wise entailment measurement in the response set $R$~\cite{williams2017broad, bowman2015large}. Following the work~\cite{kuhn2023semantic}, the employed NLI model~\footnote{off-the-shelf DeBERTa-large model} provides a three-element tuple by taking two text elements $r_i$ and $r_j$:
\begin{equation}
    [\textit{logit}_{cont}, \textit{logit}_{neut}, \textit{logit}_{ent}] = \overrightarrow{\textit{NLI}}
 (r_{i}, r_{j})
\end{equation}
The output is processed by transforming into the probability through: 
\begin{equation}
    \textbf{p} = \textit{Softmax} (\textit{logit}_{cont}, \textit{logit}_{neut}, \textit{logit}_{ent})
\end{equation}
where $\overrightarrow{p_{ent}} (r_i, r_j) = p (r_i \vdash r_j) = \textbf{p}_3$ is the entailment probability of $r_i \vdash r_j$. To here, an asymmetric entailment matrix $\mathcal{S}$ is derived for constructing the directional graph  \( \mathcal{G}_d = (V, \overrightarrow{E}) \). The \( V = R \) is the set of responses with $|V| = n$ and \( \overrightarrow{E} \) is the set of directed edges weighted primarily by the entailment probabilities.  
\begin{equation}
    \overrightarrow{E} = \{ (v_i, v_j) \mid \forall i, j \text{ weight:} p(r_i \vdash r_j) \}
\end{equation}
Thus, the adjacency matrix \( A \) of the directed graph \( \mathcal{G}_d  \) can be defined as: $A_{ij} = p(r_i \vdash r_j)$, where \( A_{ij} \) represents the weight of the directed edge from vertex \( v_i \) to vertex \( v_j \) (the vertex is indeed a corresponding response, so in a later section might use interchangeably), the $A_{ij}$ not necessarily equals to $A_{ji}$ unless the two meta responses are completely the same. The constructed $\mathcal{G}_d $ stands for directional semantic logic, different from the semantic similarity (sem) that may take the average of two entailment directions: $A_{ij, \textit{sem}} = A_{ji, \textit{sem}} = \frac{p(r_i \vdash r_j) + p(r_j \vdash r_i)}{2}$, which lacks partial of information.


\subsection{Enhance the Graph with Text Similarity}\label{sec:text}
Based on the constructed directed entailment graph, it is feasible to incorporate the text similarity to enrich the information in the graph. We consider another matrix: the text similarity matrix \( \mathcal{T} \), with identical size \( n \times n \) as $\mathcal{S}$, we can enrich the edges-carried information between the nodes with jointly weighted values from both the entailment and text similarity matrix.

Let \( \mathcal{S} = [s_{ij}] \) and \( \mathcal{T} = [t_{ij}] \) represent the entailment and text similarity matrix, respectively. We define the weight of the edge from node \( i \) to node \( j \) in the graph \( \mathcal{G}_d \) as:
$w_{ij} = s_{ij} + t_{ij}$. The adjacency matrix \( A_{ij} \) of the graph \( G \) can be updated with weights $w_{ij}$.

Please note that to achieve $\mathcal{T}$, the text similarity can be measured in multiple ways such as TF-IDF~\cite{aizawa2003information}, Cosine Similarity, Word Embeddings, etc. Here in this paper, since we are measuring the responses given the same question, we provide an implementation with Jaccard Similarity from set operation:
\begin{equation}
    J(r_i, r_j) = \frac{|r_i \cap r_j|}{|r_i \cup r_j|}
\end{equation}
where the $r_i$ and $r_j$ as response sentences, contain multiple phases and words serving as two sets.

\subsection{Random Walk Laplacian}

For a directed graph $\mathcal{G}_d$, the connectivity of nodes (responses) reflects the potential semantic clusters in the response set $R$, we can analyze the graph characteristics by conducting a Laplacian process to derive the eigenvalue, which reflects the dispersion of the nodes, in the given scenario, it reveals the uncertainty of the black box model that generated the response set, given certain question.

However, the current $\mathcal{G}_d$ is special for its asymmetric property, thus, the Normalized Laplacian or Symmetric Graph Laplacian, etc. are no longer suitable for the problem since they require the symmetric matrix. We innovatively propose to employ the Random Walk Laplacian which focuses on the out-degree of nodes to tackle this directional, and asymmetric issue. The out-degree matrix is calculated as:
$
\mathbf{D}_{\text{out}} = \text{diag}(d_{\text{out},1}, d_{\text{out},2}, \ldots, d_{\text{out},n})
$
, where \( d_{\text{out},i} = \sum_{j=1}^{n} a_{ij} \) is the out-degree of node \( r_i\), and $a_{ij}$ is an instance of adjacency matrix $A$ carrying the weights from $r_i$ to $r_j$.

Then, the inverse of the out-degree matrix can be calculated by
$
\mathbf{D}_{\text{out}}^{-1} = (\mathbf{D}_{\text{out}} + \epsilon \mathbf{I})^{-1}
$
, where \( \mathbf{I} \) is the identity matrix and \( \epsilon \) is a small positive constant to avoid division by zero. The \textbf{r}andom \textbf{w}alk Laplacian matrix \( \mathbf{L}_{\text{rw}} \) is then defined as:

\begin{equation}
\mathbf{L}_{\text{rw}} = \mathbf{I} - \mathbf{D}_{\text{out}}^{-1} \mathbf{A}
\end{equation}
we compute the eigenvalues of the random walk Laplacian matrix \( \mathbf{L}_{\text{rw}} \) and derive \( \lambda_k \), the eigenvalues of \( \mathbf{L}_{\text{rw}} \), where \( k = 1, 2, \ldots, n \). For details please refer to Appendix~\ref{proof:1}. The uncertainty measure \( U_{\text{EigV}} \) is then computed by
\begin{equation}\label{eq:egen}
    \mathbf{U}_{\text{EigV}}^d = \sum_{k=1}^{n} \max(0, 1 - \lambda_k)
\end{equation}
This measure captures the extent to which the eigenvalues $\lambda_k$ deviate from 1, providing a representation of the uncertainty in the language model's responses, note that for each question $q$ related response set $R_q$, $r_i \in R_q$, and $|R_q| = n$, our method derives one aggregated uncertainty value by Eq.~\ref{eq:egen}.

It is important to perform the Random Walk Laplacian on the directed graph $\mathcal{G}_{d}$ because in the directed graph, the probability of transition from node $i$ to node $j$ is defined by: $P_{i\rightarrow j} = \frac{A_{i\rightarrow j}}{\sum_{k}A_{i\rightarrow k}}$, where the $A_{i\rightarrow j}$ is the weights in the adjacency matrix and $k$ is the total amount of accessible nodes. If two responses exist $entail(r_i \vdash r_j) \neq entail(r_j \vdash r_i)$, then we have different transition probability, making a difference in profiling the response set characters: 
\begin{equation}
A_{i \to j} \neq A_{j \to i} \implies P_{i \to j} \neq P_{j \to i}
\end{equation}

This designed structure captures the non-symmetry information in the entailment probability from different directions of two responses. 

\subsection{Integrate Directional Entailment Uncertainty with Semantics Uncertainty}\label{sec:agg}

The uncertainty derived in Eq.~\ref{eq:egen} represents uncertainty from directional entailment probability and text in-consistency as introduced in Section~\ref{sec:text}. And since there exist multiple solutions for semantic uncertainty measurement~\cite{lin2023generating, kuhn2023semantic}, we propose a way to seamlessly integrate  $\mathbf{U}_{\text{EigV}}^d$ from \ours with other semantics uncertainty $\mathbf{U}^s$, thus have a multi-angle evaluation on limited response sets. 

One simplest way is to directly aggregate the two resources of $\mathbf{U}_{\text{EigV}}^{d,i}$ and $\mathbf{U}^{s,i}$ on the same response set $R_q^i$, however, when there are multiple response sets $R_q^i \in \mathcal{R}$, $i = \{1, 2, ...h\}$, direct aggregation can not guarantee the order change is caused by the uncertainties contribution: because the different uncertainty measure from $\mathbf{U}_{\text{EigV}}^{d,i}$ and $\mathbf{U}^s$ result in different scales (some are in [0, 1] and some are not bounded), the order changes after $\mathbf{U}_{\text{EigV}}^{d,i} + \mathbf{U}^s$ is probably caused by the absolute value range difference. Thus instead of working on $R_q^i$, we focus on the whole response space $\mathcal{R}$ that contains multiple questions' response sets, yielding $\mathbf{\mathcal{U}}_{\text{EigV}}^d$ and $\mathbf{\mathcal{U}}^s$, $|\mathbf{\mathcal{U}}_{\text{EigV}}^d| = |\mathbf{\mathcal{U}}^s| = h$ meaning there contains $h$ uncertainties for $h$ question-related response sets. And we perform the normalization on the distribution of two aspects of measurements by $Z$-score: $\text{Normalized}(X) = \frac{X - \mu_X}{\sigma_X}$, replace the X with $\mathbf{\mathcal{U}}_{\text{EigV}}^d$ and $\mathbf{\mathcal{U}}^s$ and we get: $\mathbf{\mathcal{\hat{U}}}_{\text{EigV}}^d$ and  $\mathbf{\mathcal{\hat{U}}}^s$. Then we can derive the $\mathbf{\mathcal{\hat{U}}} = \mathbf{\mathcal{\hat{U}}}_{\text{EigV}}^d / 2 + \mathbf{\mathcal{\hat{U}}}^s/2$, which contains both semantics and directional uncertainty, and the order change in $\mathbf{\mathcal{\hat{U}}}$ is contributed by the semantics uncertainty from $\mathbf{\mathcal{U}}^s$.




\section{Claim Based Response Augmentation}\label{sec:claim}

Sometimes, the raw responses from the language model $\mathcal{M}$ can not fully reveal its awareness of a problem due to the multiple claim points but short descriptions. In the example responses at Section~\ref{sec:discuss}, \texttt{`three students became heros'} and \texttt{`Andrew Willis, Chris Willis, Reece Gelea'} are a pair of responses that share the same potential meaning \texttt{`Andrew Willis, Chris Willis, and Reece Gelea are three students who became heros'}. The direct use of raw responses like these impairs ($\downarrow$) the True Positive rate and increases ($\uparrow$) the False Negative rate, leading to a biased evaluation.

In this section, inspired by~\cite{choi2024fact}, we propose to augment raw responses on the claims level, trying to identify the potential correct claims hidden in incomplete or vague responses. It is worth noting that, here we do not conduct fact-checking, instead, we rely on the context information to provide claim augmentation, so our task is easier and more feasible to be accomplished by other pre-trained LLMs. Specifically, the task can be formalized as: 

\textit{\textbf{Given a question $q$ and a response set $R = \{r_1, r_2, \ldots, r_n\} $, for each of the $r_i \in R$ that contains k claims $c_k$, augment each of $c_k \rightarrow c_k^{aug}$ and derive the  $r_{i}^{aug}$ with a more explicit and comprehensive description.
}}

To realize it, the key is to first identify the claim atoms in a response $r_i$, this step can be achieved with the basic understanding ability of context, we verified that Llama-3 is adequate for this task and used it in the claim extraction. Then to extend extracted claims by recalling the questions, this helps to align the claim descriptions with questions. And at last, combine the augmented claims into a more comprehensive answer $r_{i}^{aug}$ as follows:
\begin{equation}\label{eq:aug}
    r_i^{aug} = \textit{Augmentor} (c_1, c_2, ..., c_k)_{c_k \leftarrow r_i}
\end{equation}
where the $\leftarrow$ here is interpreted as claim $c_k$ originates \textbf{from} $r_i$. The response level $\textit{Augmentor}$ conducts two steps: 
\textbf{First}, it extends the claims with an $\textit{Extender}$ that takes into the current claim $c_k$ and question $q$ as following: 
\begin{equation}\label{eq:extend}
    c_k^{aug} = \textit{Extender}(c_k, q)
\end{equation}
the task at Eq.~\ref{eq:extend} is simple because it is generating the sequence based on existing input content, so it can be fulfilled by other general language models such as Llama-3~\footnote{\url{https://github.com/meta-llama/llama3}} (used in this paper) with necessary prompt.
\textbf{Second}, it contacts ($\oplus$) all of the augmented claims to form the $r_i^{aug}$: 
\begin{equation}
    r_i^{aug} =  \oplus \{c_1^{aug}, c_2^{aug}, ..., c_k^{aug}\}
\end{equation}
The eventual evaluation set $R^{aug}$ can be achieved by traversing all of the $r_i \in R$. 

It is worth noting that the original low-quality response $r^{\times}$ should be kept unchanged from the augmentation process to preserve the original error generated by $\mathcal{M}$ as a part of evaluation evidence, we collect these responses by regular expression as implemented in the code~\footnote{code will be released after publication and is available under request for now.}. 


In this paper, the augmentation is conducted following the above procedure, and the \ours takes the $R^{aug}$ as the eventual input for uncertainty evaluation.



\section{Experiment}
In this section, we design experiments to empirically demonstrate the effectiveness of our proposal in uncertainty evaluation for LLMs. Please note that if without extra declarition, \ours mean the directional entailment uncertainty on augmented response sets. We intend to investigate the following research questions: 

\textbf{RQ1}: Can \ours improve the uncertainty evaluation layered on existing methods that have no consideration of directional entailment logic? 

\textbf{RQ2}: Is claim level augmentation helping a more robust evaluation?

\textbf{RQ3}: How does each module of our proposed method contribute to the final uncertainty quantification? An ablation study for \ours.

\begin{figure*}[ht!]
    \centering
    \includegraphics[width=0.9\textwidth]{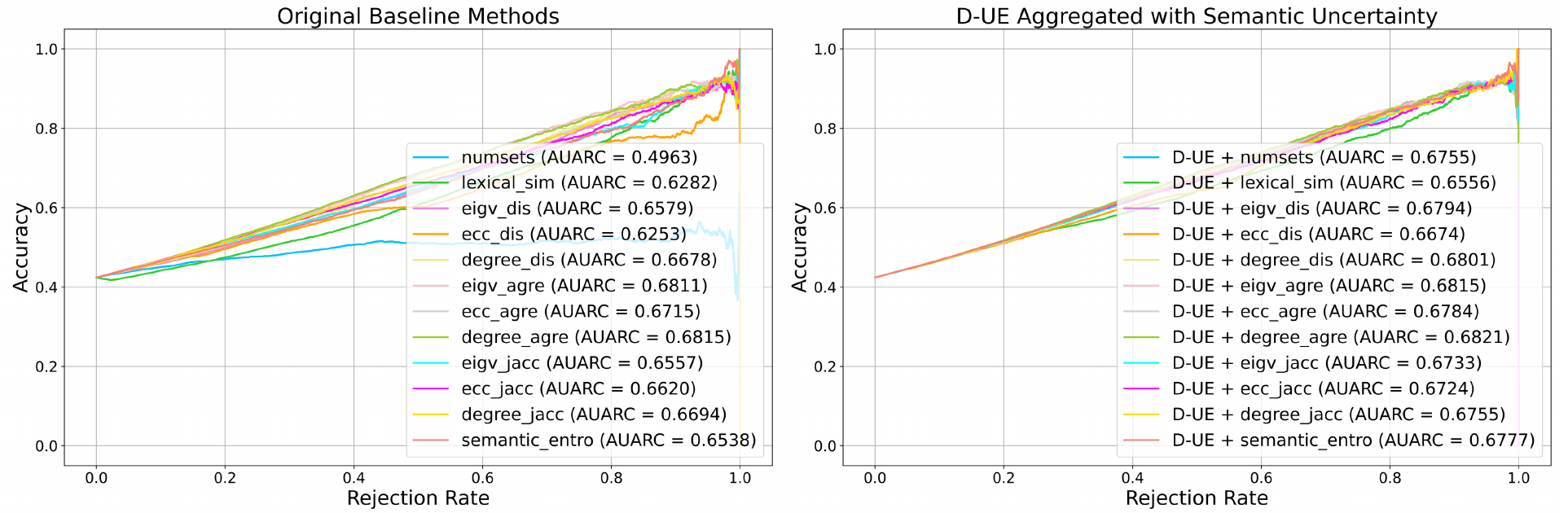}
    \caption{The comparison between \ours and baseline method on AUARC, we conducted \ours that aggregated the directional entailment uncertainty with each of the semantic measures, the evaluation improves on \texttt{Coqa} dataset.}
    \label{fig:fig2}
\end{figure*}
\subsection{Experiment Setups}

In this paper, we explored Llama3-8b for simple tasks such as claims extraction introduced in Section~\ref{sec:claim}. and question-based atom-claim augmentation to complete Eq.~\ref{eq:extend}. Each experiment using \textit{NLI} model uses a calibrated temperature as 3. All of the experiment is supported by Ubuntu on 13th Gen Intel(R) Core(TM) i9-13900KF, with NVIDIA GeForce RTX 4090.

\paragraph{Datasets}
We adopt the two classic (QA) datasets Coqa~\cite{reddy2019coqa} (7,983 questions), TriviaQA~\cite{joshi2017triviaqa} (9,960 questions), and another especially long question answer dataset NLQuAD~\cite{soleimani2021nlquad} (3,024 questions), which is more challenging and include more claims in one response.





\paragraph{Evaluation Metrics and Process}

As discussed in the paper~\cite{lin2023generating}, there exists the limitation of commonly adopted AUROC that it is very sensitive to imbalanced scenarios (likely to provide over-optimistic evaluation). Area Under Accuracy Rejection Curve (AUARC), is an alternative metric that can better reflect the evaluation performance, the calculation is shown in Appendix~\ref{app:metric}, and we use these two as a complementary evaluation indicator.
\begin{table}[h]
\centering
\resizebox{!}{0.13\textwidth}{
\begin{tabular}{@{}lc@{}}
\toprule
\textbf{Measure} & \textbf{Details} \\ 
\midrule
$U_{\textit{LexiSim}}$ & \multicolumn{1}{c}{Lexical similarity which measures the average rougeL.} \\ 
$U_{\textit{NumSet}}$ & \multicolumn{1}{c}{Multiplicity of the zero eigenvalue coincides with semantic sets.} \\ 
$U_{\textit{SE}}$ & \multicolumn{1}{c}{Semantic entropy by the entropy over  semantic sets.} \\ 
$U_{\textit{Eigv}}(Dis)$ & \multicolumn{1}{c}{Spectral eigenvalue on the disagreement.} \\ 
$U_{\textit{Ecc}}(Dis)$ & \multicolumn{1}{c}{Average distance from center in responses' disagreement.} \\ 
$U_{\textit{Degree}}(Dis)$ & \multicolumn{1}{c}{Degree of disagreement Matrix.} \\ 
$U_{\textit{Eigv}}(Agre)$ & \multicolumn{1}{c}{Spectral eigenvalue on the agreement.} \\ 
$U_{\textit{Ecc}}(Agre)$ & \multicolumn{1}{c}{Average distance from center in responses' agreement.} \\ 
$U_{\textit{Degree}}(Agre)$ & \multicolumn{1}{c}{Degree Matrix of agreement Matrix.} \\ 
$U_{\textit{Eigv}}(Jacc)$ & \multicolumn{1}{c}{Spectral eigenvalue on the Jaccard similarity.} \\ 
$U_{\textit{Ecc}}(Jacc)$ & \multicolumn{1}{c}{Average distance from center in responses' Jaccard measure.} \\ 
$U_{\textit{Degree}}(Jacc)$ & \multicolumn{1}{c}{Degree Matrix of Jaccard similarity.} \\ 
\bottomrule
\end{tabular}}
\caption{The baseline methods and explanations}
\label{tab:baslines}

\end{table}

In order to evaluate the performance of an `evaluator', we first need to know the correctness of responses to questions, then evaluate how well the evaluator's output uncertainty reflects the correctness situation (say, given a question, the more uncertainty model is, the more likely it make mistakes and achieve low accuracy). In this paper, we adopt the \texttt{GPT3.5-turbo} to produce the correctness score from 0 to 1, for details, please refer to Appenix~\ref{app:correctness}.




\paragraph{Baseline methods} 
We compare 12 baseline methods including: 
$U_{\textit{LexiSim}}$, $U_{\textit{NumSet}}$, $U_{\textit{SE}}$~\cite{kuhn2023semantic}, and similar method $\textit{Eingvalue}$-based, $\textit{eccentricity}$-based and $\textit{degree}$-based method over three characteristics: $\textit{disagreement}$, $\textit{agreement}$ and $\textit{Jaccard}$~\cite{lin2023generating} in their similarity matrix construction.
A detailed explanation is included in the following Table~\ref{tab:baslines}.

\subsection{Experiment Result and Analysis}
\begin{figure}[h!]
    \centering
    \includegraphics[width=0.50\textwidth]{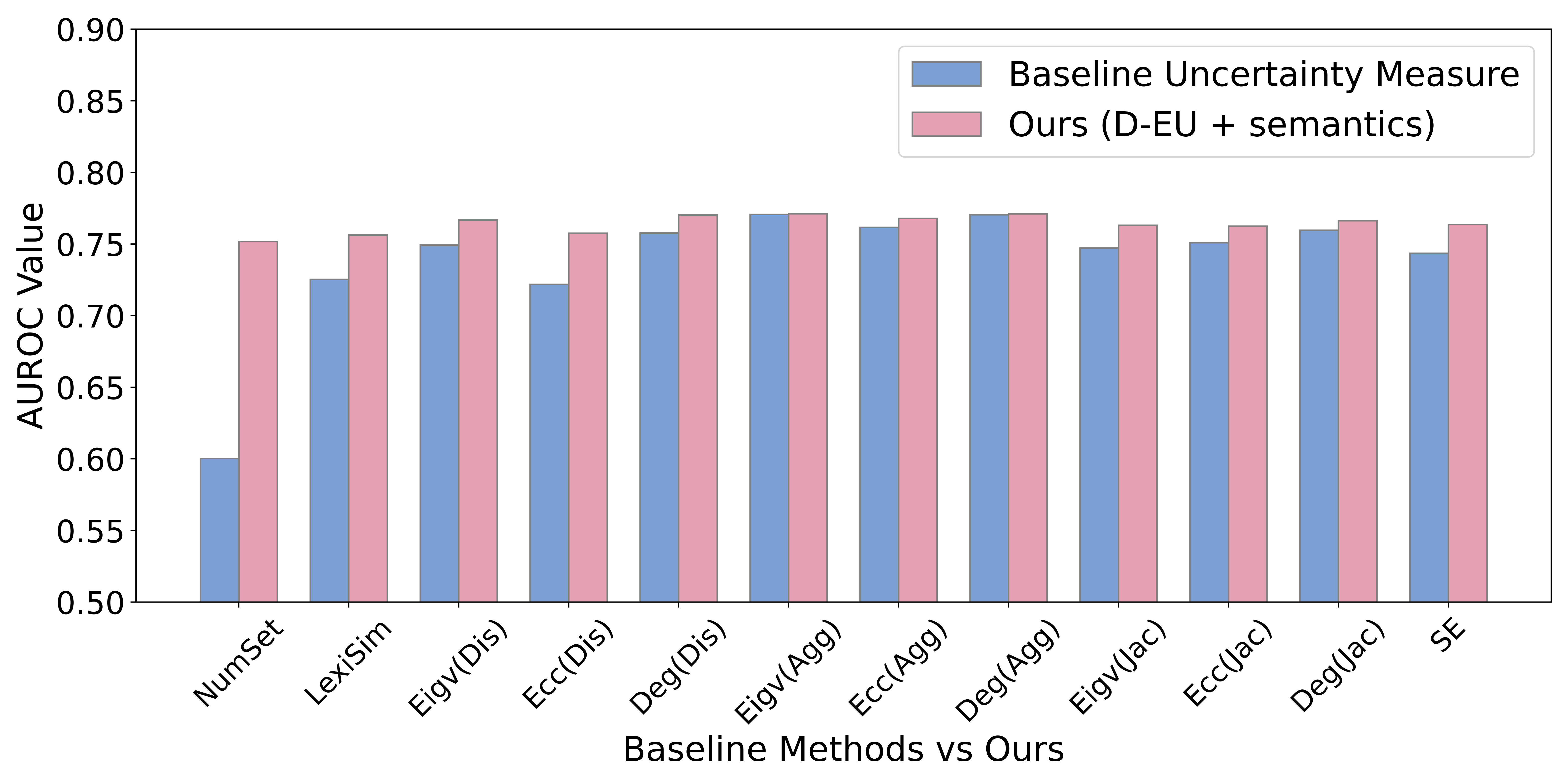}
    \caption{The comparison between \ours   and baseline method. The figure shows the evaluation from the metric of AUROC, we conducted \ours and aggregated the directional entailment uncertainty with each of the semantic measures, and the evaluation consistently improves on \texttt{Coqa} dataset.}
    \label{fig:fig1}
\end{figure}
In this section, we will discuss each of the research questions and the analysis of the proposed methods' performance.

\textbf{RQ1}: We have constructed experiments on three datasets across 12 baseline methods, and verify that the implementation with \ours + semantics uncertainty performs consistently better than most of the baseline methods. As shown in Fig.~\ref{fig:fig1}, each bar represents the area below the AUROC curve, for each of the $x$ labels, e.g, $\textit{NumSet}$, the blue color is the baseline method's performance and pink color shows the directional entailment enhanced performance on specific baseline semantic uncertainty evaluation.

The performance evaluated by AUARC is shown side by side in Fig.~\ref{fig:fig2}. The left one shows the baseline methods' performance. On the right side, our method \ours improves all of the methods' performance, which means that the directional logic information is neglected from previous methods and can be further mined by \ours. Some methods such as \texttt{numset}, \texttt{lexical\_sim} and \texttt{semanticEntropy}, could be improved a lot because they solely consider semantics similarity during the computation. On the other hand, methods like \texttt{eigv(Agre)} and \texttt{degree}-based perform smaller improvements because they also rely on the graph structure to detect the connectivity, which might consider the degree in the UQ process. But the major difference is that \ours formally defined a directed graph and conducted Random Walk Laplacian with seasonable theory approval that relaxes the symmetric requirement, and thus could be used with more flexibility. We also provide another set of experiments conducted on GPT3.5's responses on \texttt{Coqa} dataset, as shown in Table~\ref{tab:performance1}, our method outperforms all of the baseline methods. 

\begin{table}[h!]
    \centering
    \resizebox{0.44\textwidth}{!}{
    \begin{tabular}{lcccc}
        \toprule
        \multicolumn{5}{c}{Coqa (GPT3.5)} \\
        \midrule 
        & \multicolumn{2}{c}{AUARC} & \multicolumn{2}{c}{AUROC} \\
        \cmidrule(lr){2-3} \cmidrule(lr){4-5}
        Baselines & Previous & Ours & Previous & Ours  \\
        \midrule
        NumSet & 0.4250	& \cellcolor{mycustomcolor}0.5605 &0.5095	&\cellcolor{mycustomcolor}0.6660 \\
        LexiSim & 0.5001 & \cellcolor{mycustomcolor}0.5467 &0.6042& \cellcolor{mycustomcolor}0.6471\\
        Eigv(Dis) & 0.5271& \cellcolor{mycustomcolor}{0.5574} 
        & 0.6652 & \cellcolor{mycustomcolor}{0.6733} \\
        Ecc(Dis) & 0.4837 & \cellcolor{mycustomcolor}{0.5603}	& 0.5736 & \cellcolor{mycustomcolor}{0.6675} \\
        Degree(Dis) & 0.5320 & \cellcolor{mycustomcolor}{0.5579} &0.6654 & \cellcolor{mycustomcolor}{0.6736} \\
        Eigv(Agre) & 0.5355 & \cellcolor{mycustomcolor}{0.5626} &0.6769 & \cellcolor{mycustomcolor}{0.6807} \\
        Ecc(Agre) & 0.5295& \cellcolor{mycustomcolor}{0.5620} &0.6669 & \cellcolor{mycustomcolor}{0.6766}\\
        Degree(Agre) & 0.5367 & \cellcolor{mycustomcolor}{0.5615} &	0.6764 & \cellcolor{mycustomcolor}{0.6800}  \\
        Eigv(Jacc) &0.5179 & \cellcolor{mycustomcolor}{0.5579} &0.6463&\cellcolor{mycustomcolor}{0.6692}\\
        Ecc(Jacc) & 0.5173 & \cellcolor{mycustomcolor}{0.5550} &0.6544	& \cellcolor{mycustomcolor}{0.6694}\\
        Degree(Jacc) &0.5252 & \cellcolor{mycustomcolor}{0.5560} & 0.6535	& \cellcolor{mycustomcolor}{0.6693} \\
        \bottomrule
    \end{tabular}}
    \caption{Performance for Coqa under GPT3.5}
    \vspace{-2mm}
    \label{tab:performance1}
\end{table}

\textbf{RQ2}: In order to understand how claim level augmentation helps with a better understanding of the potential relationships between responses, we conducted a case study on an example response set, which is generated by llama2-13b.

 \begin{figure}[h!]
    \centering
    \includegraphics[width=0.5\textwidth]{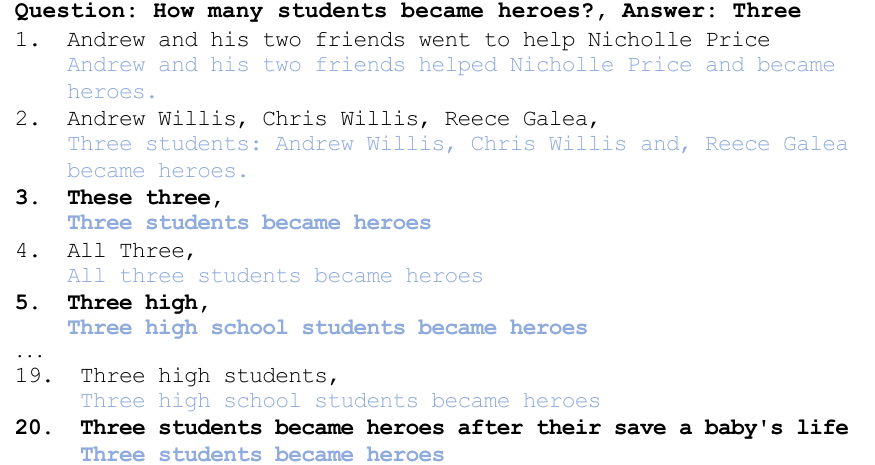}
    \label{fig:rq2}
    \vspace{-0.5cm}
\end{figure}

\begin{figure}[h!]
    \centering
    \includegraphics[width=0.50\textwidth]{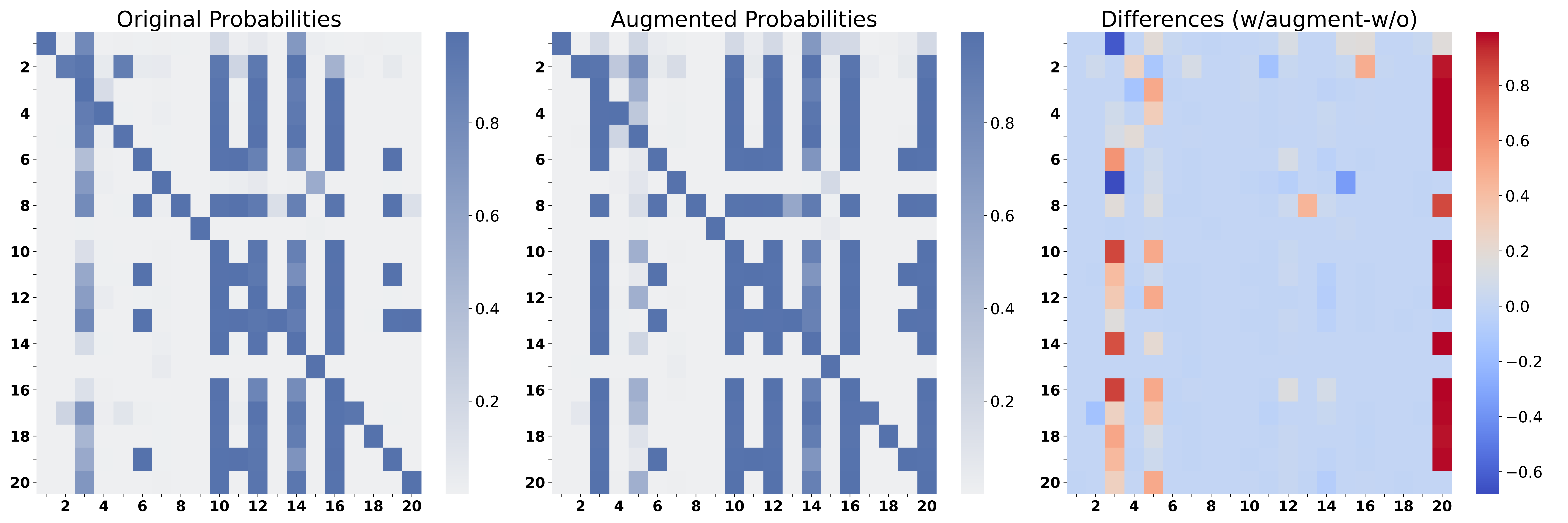}
    \caption{The entailment probability map}
    \label{fig:rq2_heatmap}
\end{figure}

As shown in this question-answer set example, which is an answer set taken by the responses made by Llama2-13b in \texttt{Coqa}
dataset, we found that, due to the quality and stability of language models, they might generate abbreviated or vague responses such as \texttt{3.`These three'} and \texttt{5.`Three high'}, even though these responses cover the key idea of the true answer, but due to the incompleteness of the sentence, this brings challenges to the uncertainty evaluation, especially for black-box evaluation that can only build upon the consistency among the responses. It is hard to identify the relationship between a sentence if the claim/meaning is not stated thoroughly. The blue color fonts show the augmented results based on our designed \texttt{Extender} (E.q.~\ref{eq:extend}), the detailed prompt will be present in the Appendix. For sentence \texttt{3.} and \texttt{5.}, these claims are completed with the intention of the questions and are easier to reflect the consistency from the meaning. As shown in Fig.~\ref{fig:rq2_heatmap}, which are the heatmaps showing the probability of entailment with direction $P(X \vdash Y)$. On the upper part of the left side is the $P_{R^{raw}}(X \vdash Y)$ from the original response set $R^{raw}$ and the lower part is after the augmentation $P_{R^{aug}}(X \vdash Y)$.
From the comparison of the original and augmented probability graph, we can observe that the probability for those in-completed sentences to entail other responses is very low, even though under the case that these sentences include the correct answer, but after the effective augmentation, we could find that the probability increases (e.g., \texttt{3.} and \texttt{5.}), indicating the potential relationship is discovered. The right side is the residual map which is calculated by subtraction from (with) augmentation to (without) augmentation, the red means finding the stronger entailment relationship after augmentation, and the blue means mitigating the original entailment probability.

\begin{figure}[ht!]
    \centering
    \includegraphics[width=0.50\textwidth]{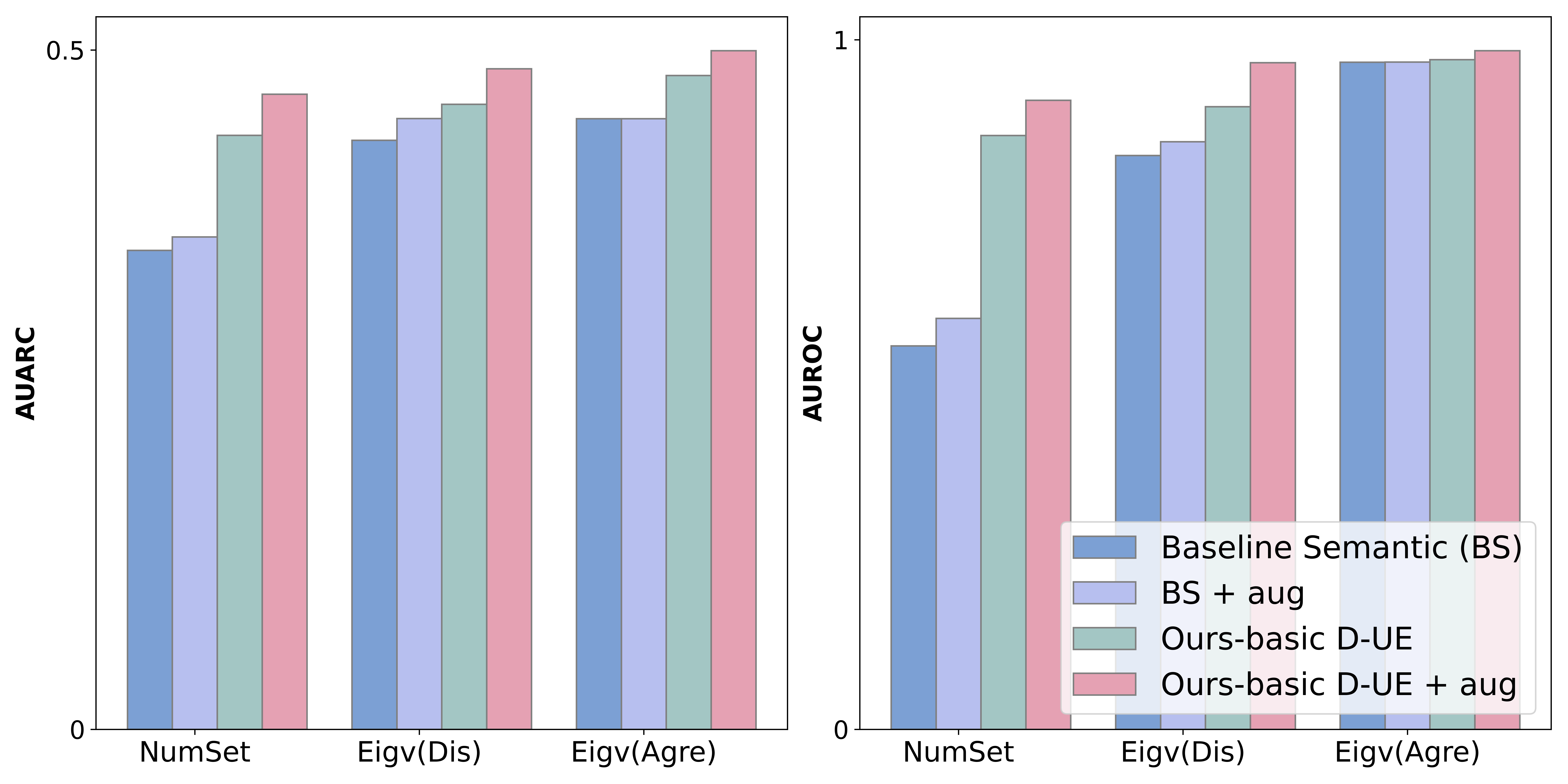}
    \caption{The comparison between baseline semantic uncertainty, \ours+semantics (no-augmentation) and full \ours+semantics}
    \label{fig:no_aug}
\end{figure}

\textbf{RQ3}: 
We conducted ablation experiments to understand the contribution of the directional entailment uncertainty measure, the claim-based augmentation. Due to the limited page, three representative baselines on two metrics are shown in Fig.~\ref{fig:no_aug}. We could observe from the baseline of \texttt{NumSet} that, the basic \texttt{NumSet} uncertainty measurement is sensitive to the augmentation by showing improvement in the augmented version (purple bar) over the basic version (blue one), both on AUROC and AUARC. But compared to the improvement brought by claim augmentation, the `Ours' method(\ours) makes a larger contribution to the general evaluation performance. This attribute to the advantage of directional entailment logic is mined with Random Walk Laplacian on the directional graph. We leave further exploration for the future on how to better combine the semantics uncertainty and directional entailment-based uncertainty and we believe there is still a potential space for improvement based on the current proposed direction.

\section{Conclusion}

In this paper, we discovered the two challenges of existing uncertainty quantification methods for LLMs: the omission of directional logic in semantic meanings and the low-quality / vague response sets that bring difficulty in uncovering the actual correct answers. We proposed two solutions to tackle the above challenges: A. we formally define a directional entailment graph encapsulating the direction logic and enhance it with text similarity, then innovatively propose to conduct Random Walk Laplacian to find the eigenvalue indicating the uncertainty in response graph structure. B. we propose a claim-based augmentation method that helps with understanding the `real' faithfulness of a model's responses. These two methods improved the current existing UQ methods and provided a better insight into how trustworthy a model is. We hope the exploration of this work could raise other researchers' interest from another aspect of understanding the uncertainty in Large Language Models and comprehending the NLG trustworthiness.

\section{Limiatations}
Even though this work innovatively proposes a directed graph and an augmentation method for the LLM's uncertainty quantification, the authors believe it is still important to explore more on how to combine the semantics and directional logic uncertainty in a theoretically orthogonal way. This work was only able to testify to the evaluation tasks on LLama2-13b and ChatGPT 3.5, with the fast-growing speed of the Large Language Model family, more models would be feasible to test and understand their responses uncertainty to specific questions.




\bibliography{custom}
\clearpage
\appendix

\section{Solve the Eigenvalue for Random Walk Laplacian}\label{proof:1}

From the definition that \[
L_{\text{rw}} = I - D_{\text{out}}^{-1} A
\]
To find the eigenvalues \( \lambda \) and eigenvectors \( \mathbf{v} \) of \( L_{\text{rw}} \) is to solve:

\[
\begin{aligned}
L_{\text{rw}} \mathbf{v} &= \lambda \mathbf{v} \\
(I - D_{\text{out}}^{-1} A) \mathbf{v} &= \lambda \mathbf{v} \\
\mathbf{v} - D_{\text{out}}^{-1} A \mathbf{v} &= \lambda \mathbf{v} \\
\mathbf{v} &= \lambda \mathbf{v} + D_{\text{out}}^{-1} A \mathbf{v} \\
(I - \lambda) \mathbf{v} &= D_{\text{out}}^{-1} A \mathbf{v} \\
D_{\text{out}} (I - \lambda) \mathbf{v} &= A \mathbf{v} \\
(D_{\text{out}} - \lambda D_{\text{out}}) \mathbf{v} &= A \mathbf{v}
\end{aligned}
\]
given this, we can transform the eigenvalue problem of \(L_{\text{rw}}\) into a form involving the matrix \(A\) and \(D_{\text{out}}\). This step simplifies the problem into the standard eigenvalue problem, specifically:
\[
\det(L_{\text{rw}} - \lambda I) = 0
\]
By applying the definition of \(L_{\text{rw}}\):
\[
\det(I - D_{\text{out}}^{-1} A - \lambda I) = 0
\]

Then the eigenvalues \(\lambda\) can be solved by solving for the roots of this equation.

\section{The Evaluation Metric}\label{app:metric}
Apart from the commonly adopted AUROC that computes the area under the ROC curve, we also employ AUARC for this paper as an evaluation metric for a more comprehensive and sensitive analysis.
The AUARC is calculated by plotting the accuracy of accepted predictions against the rejection rate, and then computing the area under this curve. 

Given that \( s_i \) is the score of the \(i\)-th prediction. \( a_i \) as the accuracy of the \(i\)-th prediction (1 if correct, 0 if incorrect). \( n \) as the total number of predictions. We first sort the scores and corresponding accuracies:
$ \{(s_i, a_i)\}_{i=1}^n \rightarrow \{(s_{(i)}, a_{(i)})\}_{i=1}^n $

For each \( i \) from 0 to \( n \):
the rejection rate \( R_i \) is:
$R_i = \frac{i+1}{n}$ and the accuracy of the accepted predictions \( A_i \) is:
$A_i = \frac{\sum_{j=i+1}^n \mathbb{\text{1}}(a_{(j)} \geq \alpha)}{n - (i+1)}$ where $ \mathbb{\text{1}}(\cdot) $ is the indicator function, and $\alpha$ is the threshold.
The area under the curve (AUARC) is calculated by the trapezoidal rule:
\begin{align*}
\text{AUARC} &= \int_0^1 A(R) \, dR \\
&\approx \sum_{i=0}^{n-1} \frac{A_i + A_{i+1}}{2} (R_{i+1} - R_i)
\end{align*}
where \( A_i \) is the accuracy at the \(i\)-th step and \( R_i \) is the rejection rate at the \(i\)-th step.
\section{Evaluation and Groundtruth Correctness}\label{app:correctness}
Following~\cite{lin2023generating}, in our evaluation, the responses with a score $>$ 0.7 are taken as correct answers, and the human verification is applied to the correctness of the auto-generated judgment by \texttt{GPT3.5-turbo} and the accuracy is about 0.95. With the ground truth correctness obtained by auto-evaluation, we can perform the evaluation on the evaluator either by AUROC or AUARC to detect how much the uncertainty quantification aligns with the correctness situation and derive the area below the ROC curve (the larger, the better) can be seen as the quality of an evaluation (UQ) method.

\section{More Experiment Results}
This section will include more details about the extra experimental results on other datasets and pair-wise comparisons.  
\subsection{The Results for Other Datasets from RQ1}
As shown in Table 3. and Table 4, \ours performs consistently better than most of the baseline methods. This indicates that our proposed solution is universally applicable to both white-box and black-box uncertainty evaluations that previously relied on the semantics information. Our method is a transferrable method to apply to any existing method that lacks the directional entailment. 
\begin{table*}[h!]
    \centering
    \resizebox{0.99\textwidth}{!}{
    \begin{tabular}{lcccc|lcccc}
        \toprule
        \multicolumn{5}{c|}{Trivia (Llama2)} & \multicolumn{5}{c}{NLQUAD (Llama2)} \\
        \cmidrule(lr){1-5} \cmidrule(lr){6-10}
        & \multicolumn{2}{c}{AUARC} & \multicolumn{2}{c|}{AUROC} & & \multicolumn{2}{c}{AUARC} & \multicolumn{2}{c}{AUROC} \\
        \cmidrule(lr){2-3} \cmidrule(lr){4-5} \cmidrule(lr){7-8} \cmidrule(lr){9-10}
        Baselines & Previous & Ours & Previous & Ours & Baselines & Previous & Ours & Previous & Ours \\
        \midrule
        NumSet & 0.7459	& \cellcolor{mycustomcolor}{0.8362} & 0.8481 & \cellcolor{mycustomcolor}{0.9422} & NumSet & 0.3525 & \cellcolor{mycustomcolor}{0.4675} & 0.5561 & \cellcolor{mycustomcolor}{0.9123} \\
        LexiSim & 0.7819 & \cellcolor{mycustomcolor}{0.8324} & 0.8174	& \cellcolor{mycustomcolor}{0.9369} & LexiSim & \cellcolor{mycustomcolor}{0.5359} & 0.5277 & 0.8872 & \cellcolor{mycustomcolor}{0.9346} \\
        Eigv(Dis) & 0.8363 & \cellcolor{mycustomcolor}{0.8457} &0.9423 & \cellcolor{mycustomcolor}{0.9593} & Eigv(Dis) & 0.4336 & \cellcolor{mycustomcolor}{0.4862} & 0.8322 & \cellcolor{mycustomcolor}{0.9669} \\
        Ecc(Dis) & 0.8185 & \cellcolor{mycustomcolor}{0.8368} & 0.9160 & \cellcolor{mycustomcolor}{0.9441} & Ecc(Dis) & 0.3918 & \cellcolor{mycustomcolor}{0.4521} & 0.6640 & \cellcolor{mycustomcolor}{0.9410} \\
        Degree(Dis) & 0.8456 & \cellcolor{mycustomcolor}{0.8491}	& 0.9614 & \cellcolor{mycustomcolor}{0.9663} & Degree(Dis) & 0.4527 & \cellcolor{mycustomcolor}{0.4970} & 0.8518 & \cellcolor{mycustomcolor}{0.9679} \\
        Eigv(Agre) & 0.8452 &	\cellcolor{mycustomcolor}{0.8454}	& \cellcolor{mycustomcolor}{0.9606} & 0.9589 & Eigv(Agre) & 0.4495 & \cellcolor{mycustomcolor}{0.4995} & 0.9674 & \cellcolor{mycustomcolor}{0.9842} \\
        Ecc(Agre) & 0.8401	& \cellcolor{mycustomcolor}{0.8435}	&0.9523	& \cellcolor{mycustomcolor}{0.9556} & Ecc(Agre) & 0.4807 & \cellcolor{mycustomcolor}{0.5117} & 0.9728 & \cellcolor{mycustomcolor}{0.9844} \\
        Degree(Agre) & \cellcolor{mycustomcolor}{0.8516} &0.8488 &	\cellcolor{mycustomcolor}{0.9727} & 0.9654 & Degree(Agre) & 0.4555 & \cellcolor{mycustomcolor}{0.5001} & 0.9656 & \cellcolor{mycustomcolor}{0.9816} \\
        Eigv(Jacc) &0.8326 & \cellcolor{mycustomcolor}{0.8422} &0.9390 & \cellcolor{mycustomcolor}{0.9537} & Eigv(Jacc) & \cellcolor{mycustomcolor}{0.5268} & 0.5180 & 0.9490 & \cellcolor{mycustomcolor}{0.9643} \\
        Ecc(Jacc) & 0.8303 & \cellcolor{mycustomcolor}{0.8399} &0.9325 & \cellcolor{mycustomcolor}{0.9487} & Ecc(Jacc) & 0.4569 & \cellcolor{mycustomcolor}{0.5082} & 0.9513 & \cellcolor{mycustomcolor}{0.9784} \\
        Degree(Jacc) & 0.8430& \cellcolor{mycustomcolor}{0.8455} & 0.9531 & \cellcolor{mycustomcolor}{0.9583} & Degree(Jacc) & \cellcolor{mycustomcolor}{0.5371} & 0.5247 &  \cellcolor{mycustomcolor}{0.9973} & 0.9835 \\
        \bottomrule
    \end{tabular}}
    \caption{Performance comparison of different baselines on Coqa and NLQUAD datasets using Llama2.}
    \label{tab:performance}
\end{table*}

\subsection{The Results for Pair-wise Comparison on \texttt{Coqa} dataset}
\begin{figure*}
    \centering
    \includegraphics[width=1.0\textwidth]{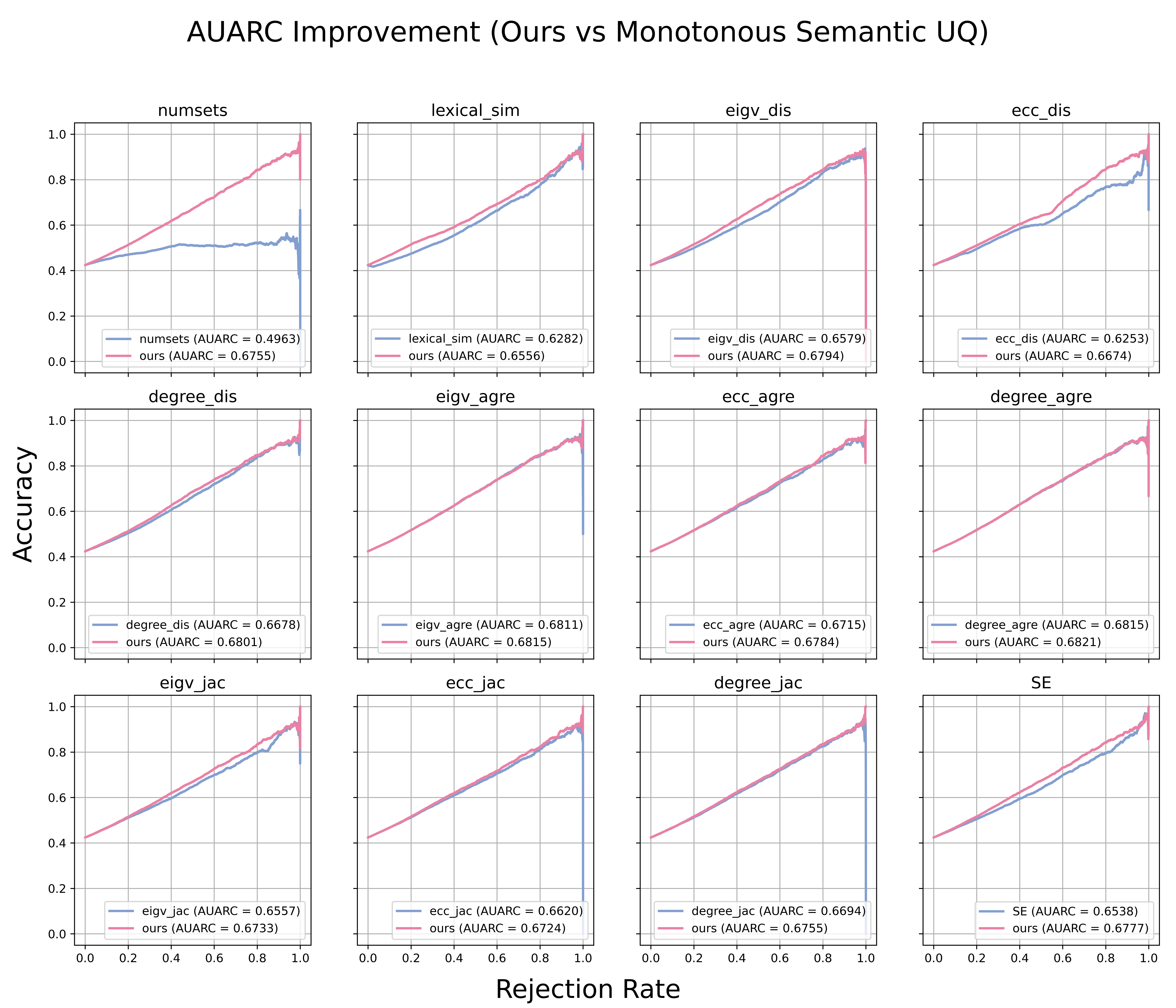}
    \caption{The AUARC improvement of our method \ours to the monotonous semantics level Uncertainty Quantification (UQ), each sub-figure demonstrates the comparison and our methods consistently perform better, note that our methods mean the method layers on the existing semantics methods and integrated with the response augmentation and directional entailment.}
    \label{fig:auarc_improve}
\end{figure*}

\begin{figure*}
    \centering
    \includegraphics[width=1.0\textwidth]{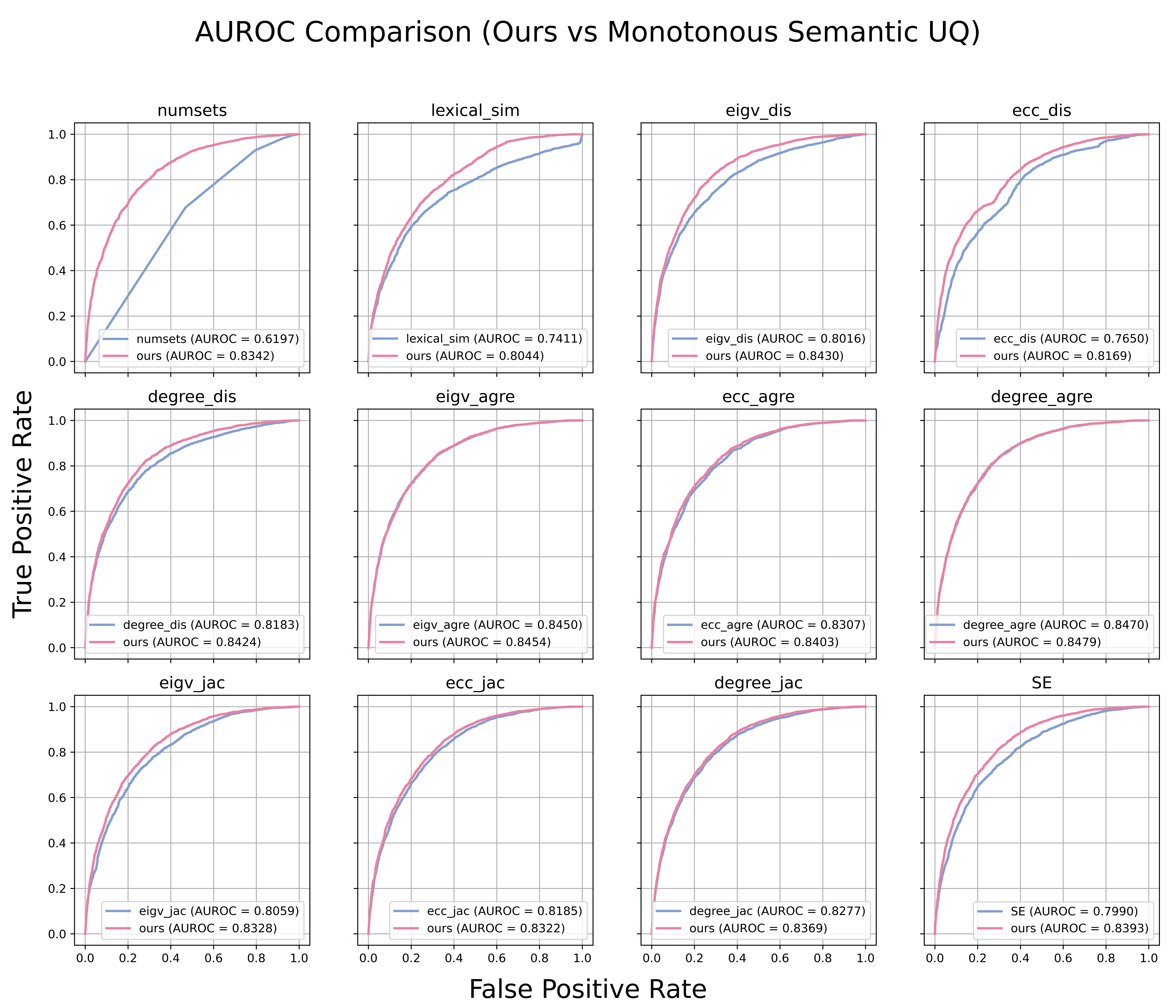}
    \caption{The AUROC improvement of our method \ours to the monotonous semantics level Uncertainty Quantification (UQ), similarly, each sub-figure demonstrates the comparison, our methods consistently perform better than solely using semantics UQ.}
    \label{fig:auroc_improve}
\end{figure*}

This subsection provides more detailed information on the pairwise comparison between \ours and traditional semantic uncertainty across AUROC and AUARC metrics. Please find in the Fig.~\ref{fig:auarc_improve} and Fig.~\ref{fig:auroc_improve}.

\begin{table}[h!]
    \centering
    \resizebox{0.48\textwidth}{!}{
    \begin{tabular}{lcccc}
        \toprule
        \multicolumn{5}{c}{Coqa (GPT3.5)} \\
        \midrule 
        & \multicolumn{2}{c}{AUARC} & \multicolumn{2}{c}{AUROC} \\
        \cmidrule(lr){2-3} \cmidrule(lr){4-5}
        Baselines & Previous & Ours & Previous & Ours  \\
        \midrule
        NumSet & 0.425	& \cellcolor{mycustomcolor}{0.5605} &0.5095	&\cellcolor{mycustomcolor}{0.6660} \\
        LexiSim & 0.5001 & \cellcolor{mycustomcolor}{0.5467} &0.6042& \cellcolor{mycustomcolor}{0.6471}\\
        Eigv(Dis) & 0.5271& \cellcolor{mycustomcolor}{0.5574} 
        & 0.6652 & \cellcolor{mycustomcolor}{0.6733} \\
        Ecc(Dis) & 0.4837 & \cellcolor{mycustomcolor}{0.5603}	& 0.5736 & \cellcolor{mycustomcolor}{0.6675} \\
        Degree(Dis) & 0.5320 & \cellcolor{mycustomcolor}{0.5579} &0.6654 & \cellcolor{mycustomcolor}{0.6736} \\
        Eigv(Agre) & 0.5355 & \cellcolor{mycustomcolor}{0.5626} &0.6769 & \cellcolor{mycustomcolor}{0.6807} \\
        Ecc(Agre) & 0.5295& \cellcolor{mycustomcolor}{0.5620} &0.6669 & \cellcolor{mycustomcolor}{0.6766}\\
        Degree(Agre) & 0.5367 & \cellcolor{mycustomcolor}{0.5615} &	0.6764 & \cellcolor{mycustomcolor}{0.6800}  \\
        Eigv(Jacc) &0.5179 & \cellcolor{mycustomcolor}{0.5579} &0.6463&\cellcolor{mycustomcolor}{0.6692}\\
        Ecc(Jacc) & 0.5173 & \cellcolor{mycustomcolor}{0.5550} &0.6544	& \cellcolor{mycustomcolor}{0.6694}\\
        Degree(Jacc) &0.5252 & \cellcolor{mycustomcolor}{0.5560} & 0.6535	& \cellcolor{mycustomcolor}{0.6693} \\
        \bottomrule
    \end{tabular}}
    \caption{Performance for Coqa under GPT3.5}
    \label{tab:performance}
\end{table}



\section{Prompt Design}
In this section, we describe the details of the prompt design for two tasks that have LLMs involved, to make sure the reproducibility of the work.
\subsection{The Prompt Used for Claim Extraction}
Firstly, we define the instructions as below:
\begin{tcolorbox}[colback=green!5,
                  colframe=black,
                  width=7.7cm,
                  arc=1mm, auto outer arc,
                  boxrule=0.05pt,
                  fontupper=\small, 
                 ]
\textcolor{red}{<<INST>><<SYS>>}

You are given a piece of text that includes knowledge claims. 
A claim is a statement that asserts something as true or false, which can be verified by humans.

\textcolor{red}{[Task] }

Your task is to accurately identify and extract every claim stated in the provided text. Then, resolve any coreference (pronouns or other referring expressions) in the claim for clarity. Each claim should be concise (less than 15 words) and self-contained. Your response MUST be a list of dictionaries. Each dictionary should contain the key "claim", which corresponds to the extracted claim (with all references resolved). You MUST only respond in the format as described below. 

\end{tcolorbox}

Then, necessary constraints and format restrictions should be applied (could vary to different LLM backbones, please modify based on the empirical exploration)

\begin{tcolorbox}[colback=purple!5,
                  colframe=black,
                  width=7.7cm,
                  arc=1mm, auto outer arc,
                  boxrule=0.05pt,
                  fontupper=\small, 
                 ]

\textcolor{red}{[Response Format]} 

[{{"claim": "Ensure that the claim is fewer than 15 words and conveys a complete idea. Resolve any coreference (pronouns or other referring expressions) in the claim for clarity." }},... ]  

\textcolor{red}{[DO NOT] }

RESPOND WITH ANYTHING ELSE. ADDING ANY OTHER EXTRA NOTES THAT VIOLATE THE RESPONSE FORMAT IS BANNED. START YOUR RESPONSE WITH '['. 

\end{tcolorbox}

Additionally, more examples are provided for few-shot learning from the inference period:
\begin{tcolorbox}[colback=gray!10,
                  colframe=black,
                  width=7.7cm,
                  arc=1mm, auto outer arc,
                  boxrule=0.05pt,
                  fontupper=\small, 
                 ]
\textcolor{red}{[Examples]:} 

\textcolor{blue}{[Text]}: 

Tomas Berdych defeated Gael Monfis 6-1, 6-4 on Saturday. The sixth seed reaches the Monte Carlo Masters final for the first time. Berdych will face either Rafael Nadal or Novak Djokovic in the final. 
\textcolor{blue}{[Response]}: 

[{{"claim": "Tomas Berdych defeated Gael Mon-fis 6-1, 6-4"}}, {{"claim": "Tomas Berdych defeated Gael Monfis 6-1, 6-4 on Saturday"}}, {{"claim": "Tomas Berdych reaches Monte Carlo Masters final"}}, {{"claim": "Tomas Berdych is the sixth-seed"}}, {{"claim": "Tomas Berdych reaches Monte Carlo Masters final for the first time"}}, {{"claim": "Berdych will face either Rafael Nadal or Novak Djokovic"}}, {{"claim": "Berdych will face either Rafael Nadal or Novak Djokovic in the final"}}]

\textcolor{blue}{[Text]}: 

Tinder only displays the last 34 photos - but users can easily see more. The firm also said it had improved its mutual friends feature. 

\textcolor{blue}{[Response]}:

[{{"claim": "Tinder only displays the last photos"}}, {{"claim": "Tinder only displays the last 34 photos"}}, {{"claim": "Tinder users can easily see more photos"}}, {{"claim": "Tinder said it had improved its feature"}}, {{"claim": "Tinder said it had improved its mutual friends feature"}}] 
\end{tcolorbox}

Given the above prompt information, we could ask for task completion to get the claims from a response (as described in the preparation step before the Eq.\ref{eq:aug}):
\begin{tcolorbox}[colback=gray!10,
                  colframe=black,
                  width=7.7cm,
                  arc=1mm, auto outer arc,
                  boxrule=0.05pt,
                  fontupper=\small, 
                 ]
 Now complete the following: 
 
 \textcolor{blue}{[Text]:} 
 
 your input text 
 
 \textcolor{blue}{[Response]:} 
 
 [/INST]'

\end{tcolorbox}

\subsection{The Prompt Used for Response Evaluation}
In this section, we provide details of prompt information used for response evaluation to get the correctness scores of the model's responses, which will be used to judge if the uncertainty evaluation result is as expected to the corresponding correctness performance.
\begin{tcolorbox}[colback=green!5,
                  colframe=black,
                  width=7.7cm,
                  arc=1mm, auto outer arc,
                  boxrule=0.05pt,
                  fontupper=\small, 
                 ]
\textcolor{red}{<<INST>><<SYS>>}

You are given a question, a reference (ground truth) answer, and an actual answer in each round of the task.

\textcolor{red}{[Task]}

Rate the level of consistency between the actual answer to the reference answer in each question. 

\end{tcolorbox}

Then we also have the value range description and reactions: 

\begin{tcolorbox}[colback=purple!5,
                  colframe=black,
                  width=7.7cm,
                  arc=1mm, auto outer arc,
                  boxrule=0.05pt,
                  fontupper=\small, 
                 ]

\textcolor{red}{[Evaluation Range] }

The evaluation value should range from 0 to 100.

\textcolor{red}{[Response Format]}  

PLEASE JUST GIVE ME A NUMBER WITHOUT ANY OTHER WORDS OR EXPLANATION.
\end{tcolorbox}

Then we can apply the few-shot learning examples to enhance the tool-LLM's understanding of its task. We provide some guidance here, and the readers could specify their demonstrations by defining a variable \texttt{few\_shots} which contains examples with a triplet of elements: (Question, Reference, Answer):

\begin{tcolorbox}[colback=gray!10,
                  colframe=black,
                  width=7.7cm,
                  arc=1mm, auto outer arc,
                  boxrule=0.05pt,
                  fontupper=\small, 
                 ]
\textcolor{red}{[Examples]:} 

Question: 
{few\_shots[0]['question']}

Reference: {few\_shots[0]['reference']}

Answer: {few\_shots[0]['answer']}
Rating: 100.

Question: {few\_shots[1]['question']}

Reference: {few\_shots[1]['reference']}

Answer: {few\_shots[1]['answer']}
Rating: 0.

\end{tcolorbox}

\subsection{The Prompt Used for Response Claim Augmentation}
This section introduces the prompt template that augments the claims extracted from the response, by reflecting on the questions being asked, the LLMs should complete the claims if any part is missing or resolve the vagueness if any sentence is found unclear.

\begin{tcolorbox}[colback=green!5,
                  colframe=black,
                  width=7.7cm,
                  arc=1mm, auto outer arc,
                  boxrule=0.05pt,
                  fontupper=\small, 
                 ]
\textcolor{red}{<<INST>><<SYS>>}

You are given two pieces of text identified as a question and response claim. A claim is a statement that asserts something as true or false, which can be verified by humans.

\textcolor{red}{[Task]}

Your task is to first understand every claim stated in the provided text. Then, augment the claims by considering the question being asked, complete the sentence if any part is missing, and resolve any coreference (pronouns or other referring expressions) in the claim for clarity. 

\end{tcolorbox}

Similarly, add the response constraint if your LLM backbone is not performing stably. After that, we could apply the agent to finish the following task by giving it the \texttt{question} and \texttt{claim} for augmentation.

\begin{tcolorbox}[colback=gray!10,
                  colframe=black,
                  width=7.7cm,
                  arc=1mm, auto outer arc,
                  boxrule=0.05pt,
                  fontupper=\small, 
                 ]
 Now complete the following: 
 
 \textcolor{blue}{[Text]:}  
 
 Question:{q},  Claim:{c} 
 
 \textcolor{blue}{[augmented claim]:} 
 
 [/INST]'

\end{tcolorbox}

Please note that the prompt performance may vary on different LLMs for completing the tasks, this prompt is testified on LLM3-8b, practitioners could tune the prompt segments if applying other backbones.


\end{document}